\documentclass[10pt,twocolumn,letterpaper]{article}

\usepackage{cvpr}              

\usepackage{graphicx}
\usepackage{amsmath}
\usepackage{amssymb}
\usepackage{booktabs}
\usepackage[accsupp]{axessibility}  

\usepackage[pagebackref,breaklinks,colorlinks]{hyperref}

\usepackage[capitalize]{cleveref}
\crefname{section}{Sec.}{Secs.}
\Crefname{section}{Section}{Sections}
\Crefname{table}{Table}{Tables}
\crefname{table}{Tab.}{Tabs.}

\def\cvprPaperID{5276} 
\def\confName{CVPR}
\def\confYear{2022}

\begin{document}

\title{Ray Priors through Reprojection: Improving Neural Radiance Fields for Novel View Extrapolation}

\author{
Jian Zhang$^{1}$\footnotemark[1]\ \ \ \ \ \
Yuanqing Zhang$^{1, 2}$\footnotemark[1]\ \ \ \ \ \
Huan Fu${^{1}}$\footnotemark[2]\ \ \ \ \ \
Xiaowei Zhou$^2$\ \ \ \ \ \
Bowen Cai$^{1}$\\
Jinchi Huang$^{1}$\ \ \ \ \ \
Rongfei Jia$^{1}$\ \ \ \ \ \
Binqiang Zhao$^{1}$\ \ \ \ \ \
Xing Tang$^{1}$\\
\\
{$^1$Tao Technology Department, Alibaba Group} \\ {$^2$State Key Lab of CAD\&CG, Zhejiang University}\\ 
}

\maketitle

\renewcommand{\thefootnote}{\fnsymbol{footnote}} 
\footnotetext[1]{These authors contribute equally to this work.} 
\footnotetext[2]{Corresponding author.} %

\begin{abstract}
Neural Radiance Fields (NeRF) \cite{mildenhall2020nerf} have emerged as a potent paradigm for representing scenes and synthesizing photo-realistic images. A main limitation of conventional NeRFs is that they often fail to produce high-quality renderings under novel viewpoints that are significantly different from the training viewpoints. In this paper, instead of exploiting few-shot image synthesis, we study the novel view extrapolation setting that (1) the training images can well describe an object, and (2) there is a notable discrepancy between the training and test viewpoints' distributions. We present RapNeRF (RAy Priors) as a solution. Our insight is that the inherent appearances of a 3D surface's arbitrary visible projections should be consistent. We thus propose a random ray casting policy that allows training unseen views using seen views. Furthermore, we show that a ray atlas pre-computed from the observed rays' viewing directions could further enhance the rendering quality for extrapolated views. A main limitation is that RapNeRF would remove the strong view-dependent effects because it leverages the multi-view consistency property.

\end{abstract}

\section{Introduction}
\label{sec:intro}
A primary target of the computer graphics community is to enable photo-realistic rendering of virtual worlds efficiently.  Physics-inspired graphics techniques well approach real-time rendering and photo-realistic imagery creation but suffer from expensive manual content generations of geometries, materials, and other aspects of scenes. The past several years have seen an explosion of interest in neural rendering \cite{tewari2020state,lombardi2019, defferedneural,liu2020neural,meshry2019neural,xu2018deep}, which models physical knowledge in deep networks to address reconstruction and rendering in a single formulation for controllable image generation. 

Leveraging neural volume rendering, a recent advance Neural Radiance Fields (NeRF \cite{mildenhall2020nerf}) learn to represent 3D scenes from images and impressively support photo-realistic image synthesis. The visual quality of the generated images is even competitive with ones produced by physically-based rendering pipelines. One of NeRF's main limitations is that it requires many images to reconstruct a scene's geometry and texture details. Thereby, several subsequent works focus on investigating few-shot or unsupervised radiance fields reconstruction \cite{yu2020pixelnerf,chen2021mvsnerf,jain2021dietnerf}. These works assume that we only observe several images of a scene. In an extreme setting, some geometries and appearances of the scene are never observed.  
\begin{figure*}[t!]
    \centering
    \includegraphics[width=0.96\textwidth]{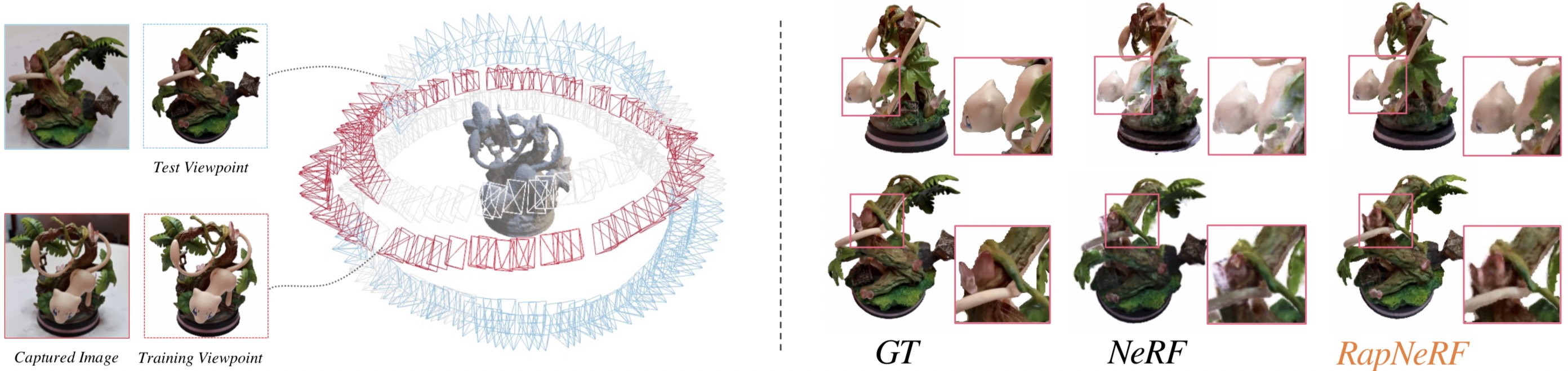}
    \caption{{\textbf{Observation \& Setting.} \emph{Left:} We study the novel view extrapolation setting that (1) the training images can well describe the objects, and (2) the test viewpoints are significantly different from the training viewpoints. We take a specific object as an example to illustrate the training (Red) and test (Blue) viewpoints in MobileObject. The viewpoints labeled in ``Gray'' are discarded. \emph{Right:} For novel view extrapolation, NeRF \cite{mildenhall2020nerf} produces images that usually contains artifacts, while RapNeRF can generate high-quality renderings.}}
    \label{fig:setting}
    \vspace{-2mm}
\end{figure*}

In this paper, we investigate NeRF from an object reconstruction perspective like \cite{reiser2021kilonerf,yu2021plenoctrees,hedman2021baking}, and restrict our focus on solid and non-transparent objects. We find that, even with enough images that can well describe an object, conventional NeRFs often fail to produce high-quality renderings for novel viewpoints that are significantly different from the training viewpoints. This observation motivates us to study the novel view extrapolation setting as explained in Figure~\ref{fig:setting}. We take inspiration from the insight that the inherent appearances of a 3D surface's arbitrary visible projections should be consistent. It has been well studied in unsupervised 3D object reconstruction and texture optimization works \cite{lin2019photometric,tulsiani2018multi,zhu2017rethinking,huang2020adversarial}. We dig into the multi-view consistency property on NeRF's formulation and present RapNeRF as a solution. 

In specific, we propose a random ray casting (RRC) policy that randomly generates rays within a cone for each training ray in an online fashion. This training strategy is simple yet effective in creating supervisions for potential unseen views using seen views. RRC relies on the target objects' rough 3D meshes (R3DMs), which can be extracted from their pre-trained NeRFs. Furthermore, we prudently rethink the tradeoff between strong view-dependent effects and multi-view consistent renderings. We show that a ray atlas (RA) computed from the training ray's viewing directions could further improve the rendering quality of extrapolated views. RapNeRF is empowered by both RRC and RA in a unified formulation. 

To study the novel view exploration setting, we resplit the synthetic training and test images of NeRF's objects to construct the Synthetic-NeRF$^{*}$ \cite{mildenhall2020nerf}. We also capture eight scenes with real objects via a mobile phone to build a MobileObject dataset. A sample is illustrated in Figure~\ref{fig:setting} (right). Experiments demonstrate the superiority of RapNeRF in synthesizing promising novel views compared to state-of-the-art approaches. We conduct various ablation studies to discuss the core components of RapNeRF. Last but not least, a major limitation of RapNeRF is that it trades some view-dependent effects for better novel view exploration performance. We provide it a remedy by studying the deferred NeRF architecture in \cite{hedman2021baking}. 

\section{Related Work}
\label{sec:related-work}

Neural Rendering (NR) bypasses mesh reconstruction to perform view synthesis of real scenes directly. It constructs an implicit scene representation from a few input images token in different viewpoints and lighting conditions \cite{tewari2020state}. This implicit scene representation can be utilized to synthesize high-quality novel images when giving some guidance. Among the NR literature, Neural Volume Rendering (NVR) supports producing photo-realistic renderings of scenes \cite{mildenhall2020nerf,bi2020deep,paschalidou2018raynet,meng2021gnerf,reiser2021kilonerf,barron2021mip,boss2020nerd,srinivasan2020nerv,raj2021pva,riegler2020free,dupont2020equivariant,wizadwongsa2021nex}. A good practice is from Neural Radiance Fields (NeRF) \cite{mildenhall2020nerf}. It proposes to represent a continuous scene as the neural radiance fields and leverages volume rendering to achieve high-quality view synthesis.  


Leveraging the success of NeRF, there are many subsequent works that have been presented for better and more efficient view synthesis \cite{lindell2020autoint,liu2020neural,yu2021plenoctrees, garbin2021fastnerf,neff2021donerf,wang2021ibrnet,lombardi2021mixture,sitzmann2020implicit,zhang2020nerf++,kosiorek2021nerf,noguchi2021neural}. For example, Neural Sparse Voxel Fields (NSVF) \cite{liu2020neural} studies a progressive training strategy to exploit sparse voxel octrees for local geometry properties modeling. The obtained model largely improve NeRF in both rendering quality and speed. Recently, PlenOctree \cite{yu2021plenoctrees} shows another milestone which realizes real-time view synthesis with preserved visual quality. Unlike NeRF's representation, PlenOctree investigates the spherical harmonic function for color computation. Other routines include deformable or dynamic scene synthesis \cite{pumarola2020d, Park20arxiv_nerfies, gafni2020dynamic, Tretschk20arxiv_NR-NeRF, rebain2020derf}, learnable camera poses \cite{wang2021nerf,yen2020inerf,lin2021barf}, and editable view synthesis \cite{xiang2021neutex, niemeyer2020giraffe}.

Several works learn few-shot view synthesis by conditioning a NeRF on image inputs \cite{yu2020pixelnerf} and exploiting the semantic consistency of multi-view features \cite{jain2021dietnerf}. In the few-shot setting, they only observe several images of a scene. Some geometries and appearances are not covered by these images. This makes the neural reconstruction process particularly challenging. In contrast, we study a relaxed novel view extrapolation setting that the training images are enough to well describe an object. There is a great concurrent work RegNeRF \cite{niemeyer2021regnerf} regularizes the geometry and appearance leverages the multi-view consistency property to obtain 3D-consistent representations.

\section{Methodology}
\label{sec:method}

Our main goal is to close the visual quality gap between interpolated and extrapolated views synthesized by NeRF \cite{mildenhall2020nerf} series. In this section, we begin with a brief review of NeRF in Sec.~\ref{subsec:NeRF}. Then, we present the proposed random ray casting (RRC) policy in Sec.~\ref{subsec:rrc} and explain the ray atlas (RA) in Sec.~\ref{subsec:rm}. RRC and RA rely on objects' rough 3D meshes, which are extracted from the pre-trained NeRFs. Finally, we present how we train RapNeRF in Sec.~\ref{subsec:train}. 

\begin{figure*}[t!]
    \centering
    \includegraphics[width=0.97\textwidth]{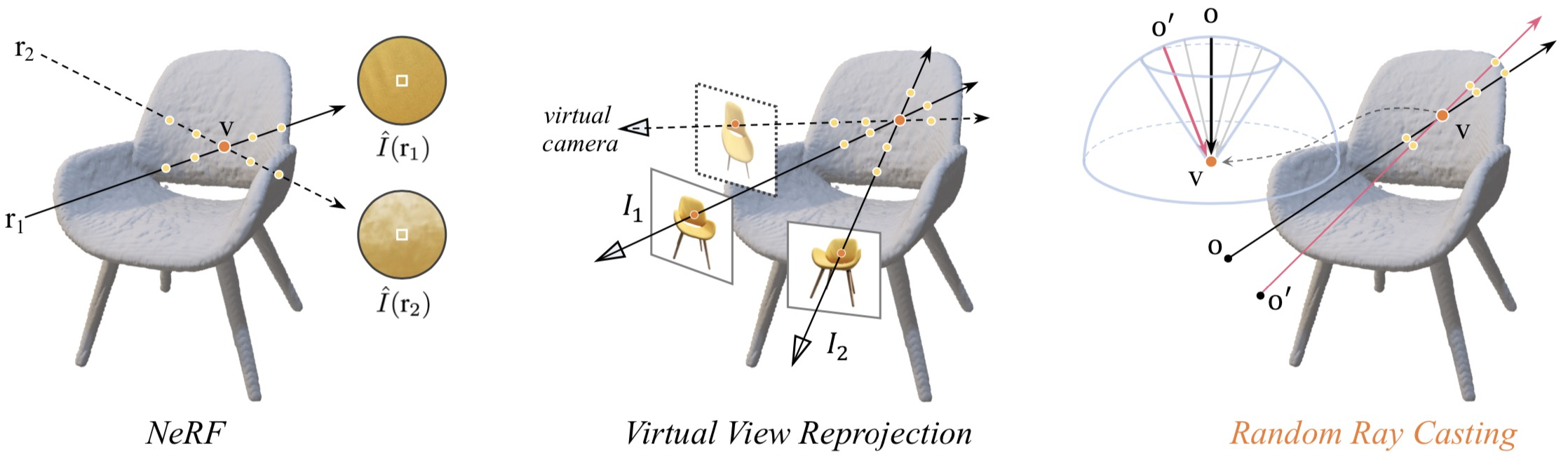}
    \caption{{\textbf{Random Ray Casting.} \emph{Left:} $\text{r}_1$ lies in the training space, and $\text{r}_2$ is distant from training rays. The radiance accumulation operation along $\text{r}_2$ is more likely to provide an adverse color estimation of $\text{v}$ compared to $\hat{I}(\text{r}_1)$. \emph{Middle}: A straightforward virtual view reprojection idea, which is inconvenient. \emph{Right}: For a specific training ray (casting from $\text{o}$ and passing through $\text{v}$), the random ray casting (RRC) policy randomly generate an unseen virtual ray (casting from $\text{o}'$ passing through $\text{v}$) within a cone, then assign it a pseudo label based on the training ray in an online manner. RRC enables training unseen rays using seen rays. See Sec.~\ref{subsec:rrc} for detailed explanations.
    }}
    \label{fig:rrc}
    \vspace{-2mm}
\end{figure*}

\subsection{Preliminaries: NeRF}
\label{subsec:NeRF}

A volume scene representation can be seen as a radiance field (or a 5D vector-valued function) that takes a 3D location $\text{x} = (x, y, z)$ and 2D viewing direction $\text{d} = (\theta, \Phi)$ as input, and output an emitted color (or radiance) $\text{c} = (r, g, b)$ and a volume density $\sigma$. NeRF \cite{mildenhall2020nerf} adopts a single MLP network to approximate the 5D function as the neural radiance fields:
\begin{equation}
    \begin{aligned}
      \sigma, \text{c} &= F(\text{d}, \text{x}).
    \end{aligned}
\end{equation}
The formulation can be further decomposed as $F_{\sigma}:\text{x} \to (\sigma, \text{f})$ and  $F_{\text{c}}:(\text{d}, \text{f}) \to \text{c}$.

To render a pixel of image $I$, NeRF casts a ray $\text{r}$ from the camera’s center of projection \text{o} along the direction \text{d} passing through the pixel. It samples $N$ points along the ray and approximate the pixel's color $\hat{I}(\text{r})$ following:
\begin{equation}
    \begin{aligned}
      \hat{I}(\text{r}) &= \sum_{i=1}^{N} T_i(1 - \text{exp}(-\sigma_i\delta_i))\text{c}_i, \\
      T_i &= \text{exp}(-\sum_{j=1}^{i-1}\sigma_j\delta_j),
    \end{aligned}
\end{equation}
where $\delta_i = t_{i+1} - t_i$ denotes the distance between two consecutive samples, $\text{c}_i$ and $\sigma_i$ are the radiance and volume density of a sample point $\text{r}(t_i) = \text{o} + t_i\text{d}$, and $T_i$ represents the accumulated transmittance from $\text{r}(t_1)$ to $\text{r}(t_i)$. In practice, $t_i$ are bounded by a predefined intervals $[t_n, t_f]$. NeRF minimizes the squared error between the rendered and true pixel colors ($\hat{I}(\text{r})$ \emph{vs.} $I(\text{r})$) to learn its MLP.

\subsection{Random Ray Casting}
\label{subsec:rrc}
We start with a further discussion of the aforementioned visual quality gap from the ray casting and neural mapping perspectives. As shown in Figure~\ref{fig:rrc} (left), $\text{r}_1$ and $\text{r}_2$ are two rays that view a 3D point $\text{v}$ in two directions, where the former lies in the training space, and the latter (a test ray) is distinct from the training rays. We may have a sense that the radiance of some samples along $\text{r}_2$ would be imprecise considering both the distribution shift and the mapping function $F_{\text{c}}:(\text{r}, \text{f}) \to \text{c}$. In further, the radiance accumulation operation along $\text{r}_2$ is more likely to provide an adverse color estimation compared to $\hat{I}(\text{r}_1)$. We can
see from the synthesized small regions around $\hat{I}(\text{r}_2)$ and $\hat{I}(\text{r}_1)$, the former contains more artifacts.

Our intuition is to create supervisions for potential unseen views by exploiting the multi-view consistency property. The property has been well studied in reprojection-based unsupervised 3D object reconstruction and texture optimization approaches \cite{yang2018learning,lin2019photometric,tulsiani2018multi,zhu2017rethinking,huang2020adversarial}. Here in the NeRF formulation, we can naively follow the pipeline of generating some virtual cameras and their views, computing the involved pixel-wise rays, finding the corresponding rays for each virtual ray that hit the same 3D surface point from the training ray pool. It likes a pseudo-label generation process for virtual rays through multi-view projection. See Figure~\ref{fig:rrc} (middle) for an illustration of this virtual view reprojection solution. In practice, the offline workflow is inconvenient.

Our random ray casting (RRC) policy allows pseudo-label assignment for randomly generated virtual rays in an online manner. Specifically, for an interested pixel in one training image $I$, we are given its viewing direction $\text{d}$, camera origin $\text{o}$, and depth value $t_\text{z}$ in the world coordinate system, and ray $\text{r} = \text{o} + t\text{d}$. Here, $t_\text{z} = \sum_{i=1}^{N}T_i(1 - \text{exp}(-\sigma_i\delta_i))t_i$ is pre-computed and stored using the pre-trained NeRF. Let $\text{v} = \text{o} + t_\text{z}\text{d}$ denote the closest 3D surface point hit by $\text{r}$. In the training phase, as shown in Figure~\ref{fig:rrc} (right), we regard $\text{v}$ as a new origin, and randomly cast a ray from $\text{v}$ within the cone whose central ray is the vector $\overrightarrow{\text{v}\text{o}} = -t_\text{z}\text{d}$. This can be easily implemented by converting $\overrightarrow{\text{v}\text{o}}$ to the spherical space, and introducing some randomness $\Delta\varphi$ and $\Delta\theta$ to $\varphi$ and $\theta$. Here, $\varphi$ and $\theta$ are the \emph{Azimuth} and \emph{Elevation} of $\overrightarrow{\text{v}\text{o}}$, respectively. $\Delta\varphi$ and $\Delta\theta$ are uniformly sampled from a pre-defined interval $[-\eta, \eta]$. With this operation, we obtain $\theta' = \theta + \Delta\theta$ and $\varphi' = \varphi + \Delta\varphi$, thus can generate a virtual ray $\text{r}'$ casting from a random origin $o'$ that also passes through $\text{v}$. Thereby, we can treat the ground truth color intensity $I(\text{r})$ as the pseudo label of $\hat{I}(\text{r}')$. 



\begin{figure}[t!]
    \centering
    \includegraphics[width=0.5\textwidth]{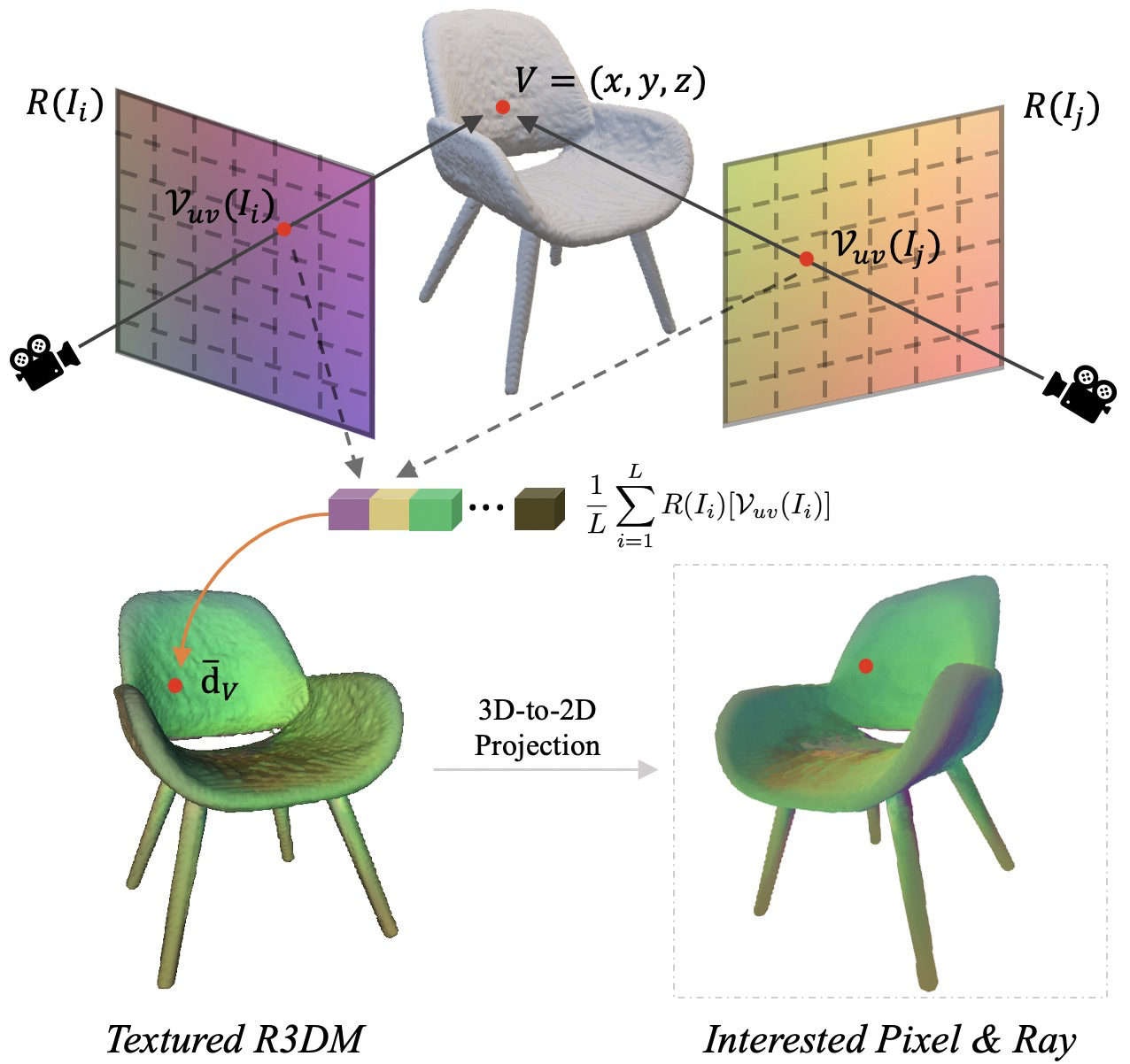}
    \caption{{\textbf{Ray Atlas.} An illustration of how we capture a ray atlas from the training rays and use it to texture a chair's rough 3D mesh (R3DM). $R(I_i)$ is the ray map of the training image $I_i$. $\mathcal{V}_{uv}(I_i)$ is the 2D position of image $I_i$ corresponding to vertex $V$. The global ray direction $\bar{\text{d}}_V$ of vertex $V$ is computed following Eqn.~\ref{eqn:rm}.
    }}
    \label{fig:rm}
    \vspace{-0.3cm}
\end{figure}

\subsection{Ray Atlas}
\label{subsec:rm}
The vanilla NeRF utilizes ``direction embedding" to encode the lighting effects of a scene. We find the scene fitting process makes the trained color prediction MLP rely heavily on the viewing direction. It is not a problem for novel view interpolation. Nevertheless, it might not be good for novel view extrapolation as there are some discrepancies between the training and test ray distributions. A naive idea is to directly remove the direction embedding (denoted as ``NeRF w/o dir"). However, we find it often produces images with artifacts such as unexpected ripple and non-smooth colors. That means the rays' viewing directions might also contribute to surface smoothing. We compute a ray atlas and show it can further enhance the rendering quality of extrapolated views while not involving more issues to interpolated views. A ray atlas is like a texture atlas, but instead, it stores a global ray direction for each 3D vertex. 

In particular, for each image (\emph{e.g.,} image $I$), we capture its rays' viewing directions for all spatial locations, resulting in a ray map $R(I)$. We extract a rough 3D mesh (R3DM) from the pre-trained NeRF, and map the ray directions to the 3D vertexes. Taking a vertex $V = (x, y, z)$ as an example, its global ray direction $\bar{\text{d}}_V$ should be expressed as:
\begin{equation}\label{eqn:rm}
    \begin{aligned}
      &\bar{\text{d}}_V = \frac{1}{L}\sum_{i=1}^L R(I_i)[\mathcal{V}_{uv}(I_i)], \\ &\mathcal{V}_{uv}(I_i) = \frac{1}{z}K\mathcal{T}_{w2c}(I_i)V, \\
    \end{aligned}
\end{equation}
where $K$ is the camera intrinsic parameter, $\mathcal{T}_{w2c}(I_i)$ is the camera to world transformation matrix of image $I_i$, $\mathcal{V}_{uv}(I_i)$ denotes the projected 2D location in image $I_i$ of vertex $V$, and $L$ represents the number of training images that contributed to the reconstruction of vertex $V$. We normalize $\bar{\text{d}}_V$ before storing it. Then, for each pixel with an arbitrary camera pose, we can capture a global ray prior $\bar{\text{d}}$ by projecting the R3DM, which is textured by the ray maps, to 2D. See Figure~\ref{fig:rm} for an illustration.

When training RapNeRF, we adopt $\bar{\text{d}}$ of the interested pixel $I(\text{r})$ to replace its $\text{d}$ in $F_c$ for its color prediction. This alternative mechanism occurs with a probability of 0.5. In our experiments, we sample points along the original ray $\text{r}$, and use $\bar{\text{d}}$ for ray embedding computation.  We find sampling points along the direction $\bar{\text{d}}$ would sometimes make the training unstable. In the test phase, the radiance $\text{c}$ of a sample $\text{x}$ is approximated as:
\begin{equation}
    \begin{aligned}
      \text{c} &= F_c(\bar{\text{d}}, F_\sigma(\text{x})). \\
    \end{aligned}
\end{equation}




\subsection{Training RapNeRF}
\label{subsec:train}
Our RapNeRF is trained in two stages. For an object to be reconstructed, we first train a NeRF \cite{mildenhall2020nerf} in $N_1$ iterations to recovery the geometry. Then, we incorporate the proposed random ray cast (RRC) policy and ray mapping (RA) approach to fine-tune the pre-trained NeRF in another $N_2$ iterations. For each iteration, RRC and RA occur with the probabilities of 0.7 and 0.5, respectively. We set $\eta$ in RRC to 
$30^{\circ}$. In both stages, we employ an additional opacity constraint \cite{niemeyer2020differentiable} to enforce the accumulated opacity (transmittance) along a ray to be 1 if the ray trace through the object regions, and 0 if the ray belongs to the backgrounds. Let $m(\text{r})$ denote the mask label (1 or 0) of a pixel ray. The opacity constraint can be expressed as:
\begin{equation}
    \begin{aligned}
      \mathcal{L}_{o} &= \sum_{\text{r}} \lvert m(\text{r}) + T_{N}(\text{r}) - 1 \rvert,
    \end{aligned}
\end{equation}
where $T_N(\text{r})$ can be seen as the ratio of light that can pass through the object. $\mathcal{L}_{o}$ could reduce some noisy volume densities around the objects' surface regions. It is worth mentioning that other NeRF works \cite{hedman2021baking,yu2021plenoctrees,liu2020neural,xiang2021neutex} also incorporate opacity regularization techniques to remove background voxels for object reconstruction. 


\setlength\tabcolsep{12pt}
\begin{table*}[t!]
\centering
\begin{tabular}{ c c c c c c c }
\toprule
 & \multicolumn{3}{c}{Synthetic-NeRF$^{*}$ \cite{mildenhall2020nerf}} & \multicolumn{3}{c}{MobileObject} \\
 \cmidrule(lr){2-4}\cmidrule(lr){5-7}
 Method & PSNR $\uparrow$ & SSIM $\uparrow$ & LPIPS $\downarrow$ & PSNR $\uparrow$ & SSIM $\uparrow$ & LPIPS $\downarrow$ \\
\midrule\midrule
NeRF \cite{mildenhall2020nerf} & 25.73 & 0.906 & 0.090 & 24.05 & 0.948 & 0.089 \\
NeRF w/o dir \cite{mildenhall2020nerf} & 26.14 & 0.918 & 0.067 & 26.69 & 0.951 & 0.053 \\
NSVF \cite{liu2020neural} & 26.37 & 0.893 & 0.088 & 22.65 & 0.899 & 0.125 \\
IDR \cite{yariv2020multiview} & 20.45 & 0.909 & 0.113 & 23.43 &	0.947 & 0.074 \\
IBRNet \cite{wang2021ibrnet} & 23.33 & 0.870 & 0.153 & 18.99 & 0.870 & 0.185 \\
PlenOctree \cite{yu2021plenoctrees} & 24.19 & 0.875 & 0.100 & 21.76 & 0.903 & 0.105 \\
SNeRG$^\dag$ \cite{hedman2021baking} & 24.68 & 0.904 & 0.074 & 26.32 & 0.952 & 0.058 \\
RapNeRF & \textbf{27.63} & \textbf{0.929} & \textbf{0.046} & \textbf{28.90} & \textbf{0.963} & \textbf{0.045} \\
RapNeRF$^\dag$ & 26.40 & 0.912 & 0.069 & 28.65 & 0.961 & 0.047 \\
\bottomrule
\end{tabular}
\caption{\textbf{Benchmark Comparisons.} RapNeRF$^\dag$ remedies the degenerated view-dependent effects of RapNeRF by incorporating RRC and RA into a deferred NeRF variant (SNeRG$^\dag$ \cite{hedman2021baking}). It further shows RRC and RA can be easily integrated into other NeRFs. Our approaches obtain best performance on datasets of both synthetic and real images. IBRNet \cite{wang2021ibrnet} and IDR \cite{yariv2020multiview} also exploit the multi-view consistency property. See Sec.~\ref{sec:limitation} for limitation discussions and Sec.~\ref{subsec:bc} for experimental details. Per-object results are reported in the supplementary.}
\label{tab:bc}
\end{table*}

\setlength\tabcolsep{4.8pt}
\begin{table}[t!]
\centering
\begin{tabular}{ c c c c c c }
\toprule
\multicolumn{3}{c}{Components} & \multicolumn{3}{c}{Metrics} \\
 \cmidrule(lr){1-3}\cmidrule(lr){4-6}
 NeRF \cite{mildenhall2020nerf} & RRC & RA &  PSNR $\uparrow$ & SSIM $\uparrow$ & LPIPS $\downarrow$ \\
\midrule\midrule
$\surd$ &  &  & 24.05 & 0.948 & 0.089 \\
$\surd$ & $\surd$  &  & 27.55 & 0.963 & 0.045 \\
$\surd$ &  & $\surd$ & 25.29 & 0.954 & 0.056 \\
$\surd$ & $\surd$ & $\surd$ & \textbf{28.90} & \textbf{0.963} & \textbf{0.045} \\
\bottomrule
\end{tabular}
\caption{\textbf{Ray Priors.} We study the performance gains ordained by the proposed random ray casting (RRC) and ray atlas (RA) approaches on the MobileObject dataset. When computing $\hat{I}(\text{r})$, RapNeRF adopts the direction embedding of $\text{d}$ (if only use RRC) or the direction embedding of $\bar{\text{d}}$ from the ray atlas (if use RA).}
\label{tab:rap}
\end{table}

\section{Experiments}
\label{sec:exp}
In this section, we conduct experiments to investigate the performance of our RapNeRF for novel view extrapolation. First, we briefly introduce the Synthetic-NeRF$^{*}$ \cite{mildenhall2020nerf} and MobileObject datasets towards our studied setting in Sec.~\ref{subsec:dataset}. Then, we make qualitative and quantitative comparisons with recent representative NeRF variants in Sec.~\ref{subsec:bc}. Finally, we perform various ablation studies based on MobileObject to discuss our method in Sec.~\ref{subsec:ablation}. We refer to the supplemental materials for more experimental results. 

\subsection{Datasets}
\label{subsec:dataset}
\noindent \textbf{Synthetic-NeRF$^{*}$.}  The original Synthetic-NeRF \cite{mildenhall2020nerf} dataset contains eight objects, where each object is rendered in a resolution of $800 \times 800$, with 100 views for training and 200 for testing. The viewpoints are sampled on the upper hemisphere or a full sphere. The ground-truth camera poses and object masks are provided. In this paper, we simply sort the cameras' locations along the $z$ axis in ascending order, and choose the first 100 images for training, and the remained 200 images for testing. 
\newline

\noindent \textbf{MobileObject.} We capture eight object-centric videos using a mobile device, where the viewpoints are on the upper space. To allow better SFM, we put the target objects into complexity scenes that are with rich textures before recording the videos. The image size is $960 \times 540$ or $540 \times 960$. For each object, we uniformly sample about 200$\sim$300 images from the video sequence and compute the camera poses using COLMAP \cite{schonberger2016structure}. The blur images are pre-removed. Then, we compute the average $z$ value of the cameras' positions, and choose its neighboured 100 images for training based on the $z$ value distance. To construct the test set, we utilize the following metric to measure the distance between a camera pose $y \in \mathcal{SO}(3)$ and the training poses $\mathcal{X}$:
\begin{equation}\label{eqn:pose-distance}
    \begin{aligned}
      \mathcal{D}_y &= \mathcal{\min}_{x \in \mathcal{SO}(3)}\{\left\| \log(x) - \log(y) \right\| \mid x \in \mathcal{X} \},
    \end{aligned}
\end{equation}
where $\mathcal{SO}(3)$ is the 3D rotation group. A large $\mathcal{D}_y$ represents a significant viewpoint discrepancy. We compute $\mathcal{D}_y$ for each remained image, and choose the ones with larger distance for testing. We select up to 60$\sim$100 test images for each object according to its video length. Other images have been filtered out.


\begin{figure*}[t!]
    \centering
    \includegraphics[width=1.0\textwidth]{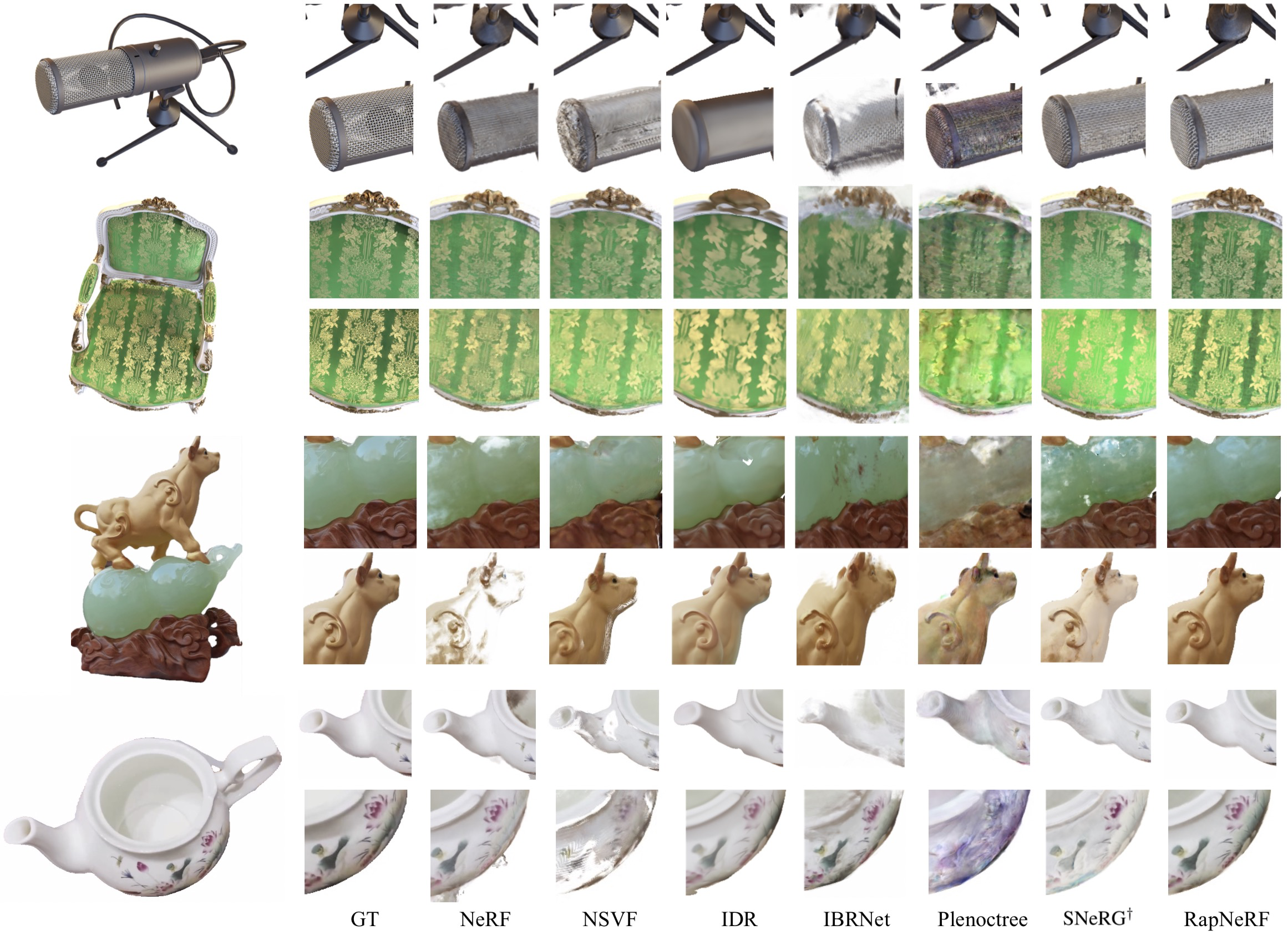}
    \caption{{\textbf{Qualitative Comparisons.} We compare RapNeRF with several recent impressive methods on Synthetic-NeRF$^*$ (Top) and MobileObject (Bottom). RapNeRF better recover the fine details of these objects. The images synthesized by other methods often contains artifacts. We refer to the supplemental materials for more qualitative comparisons. Zoom in for a better view.
    }}
    \label{fig:comparisons}
\end{figure*}

\subsection{Benchmark Comparisons}
\label{subsec:bc}
We make comparisons with NeRF \cite{mildenhall2020nerf} and its recent represent variants, including NSVF \cite{liu2020neural}, IBRNet \cite{wang2021ibrnet}, PlenOctree \cite{yu2021plenoctrees}, and SNeRG \cite{hedman2021baking}. We perform per-object fine-tuning for IBRNet using their released model pre-trained on a large database. PlenOctree here means its NeRF-SH version. We slightly reformulate the deferred NeRF architecture in SNeRG by predicting a specular color for each sampled point along a ray from $\text{f}$ and $\text{d}$. We have not used its sparse radiance grid data structure as it imposes a quality loss of about 2dB. We also examine IDR \cite{yariv2020multiview}, an impressive 3D reconstruction work that exploits the multi-view consistency property leveraging differentiable rendering. We optimize NeRF in $N_1 + N_2$ iterations for a fair comparison, since RapNeRF contains a pre-training stage. For other methods, we train them longer to ensure their models are converged. 

We use PSNR, SSIM \cite{wang2004image}, and LPIPS \cite{zhang2018unreasonable} to measure the rendering quality. As reported in Table~\ref{tab:bc}, RapNeRF obtains the best performance on all metrics for novel view extrapolation. It is interesting to see the vanilla NeRF achieves higher PSNR than NSVF on the real MobielObject dataset and vice verve on Synthetic-NeRF$^{\star}$. As discussed in the NSVF paper \cite{liu2020neural}, it shows fewer tolerances to camera pose errors than NeRF. IBRNet's performance drops much on real scenes (18.99 \emph{vs}. 23.33 for synthetic scenes). The main reason is that the pose distance between training and test views of MobileObject's scenes is large. IBRNet and a similar work MVSNeRF \cite{chen2021mvsnerf} can only well reconstruct the scene content that has been seen by the reference views (a small scene frustum). IDR here yields a mean PSNR of 23.43 on real scenes. We notice that it performs similarly on the DTU dataset \cite{jensen2014large} (23.20 as reported in its paper). That means, though IDR is good at recovering surfaces, but is not as great as NeRF for photo-realistic view synthesis. We also show some qualitative results in Figure~\ref{fig:comparisons}. Other methods often produce renderings that contain artifacts and distortions, while RapNeRF can generate images with great visual quality.

\subsection{Ablation Studies}
\label{subsec:ablation}

\begin{figure}[t!]
    \centering
    \includegraphics[width=0.49\textwidth]{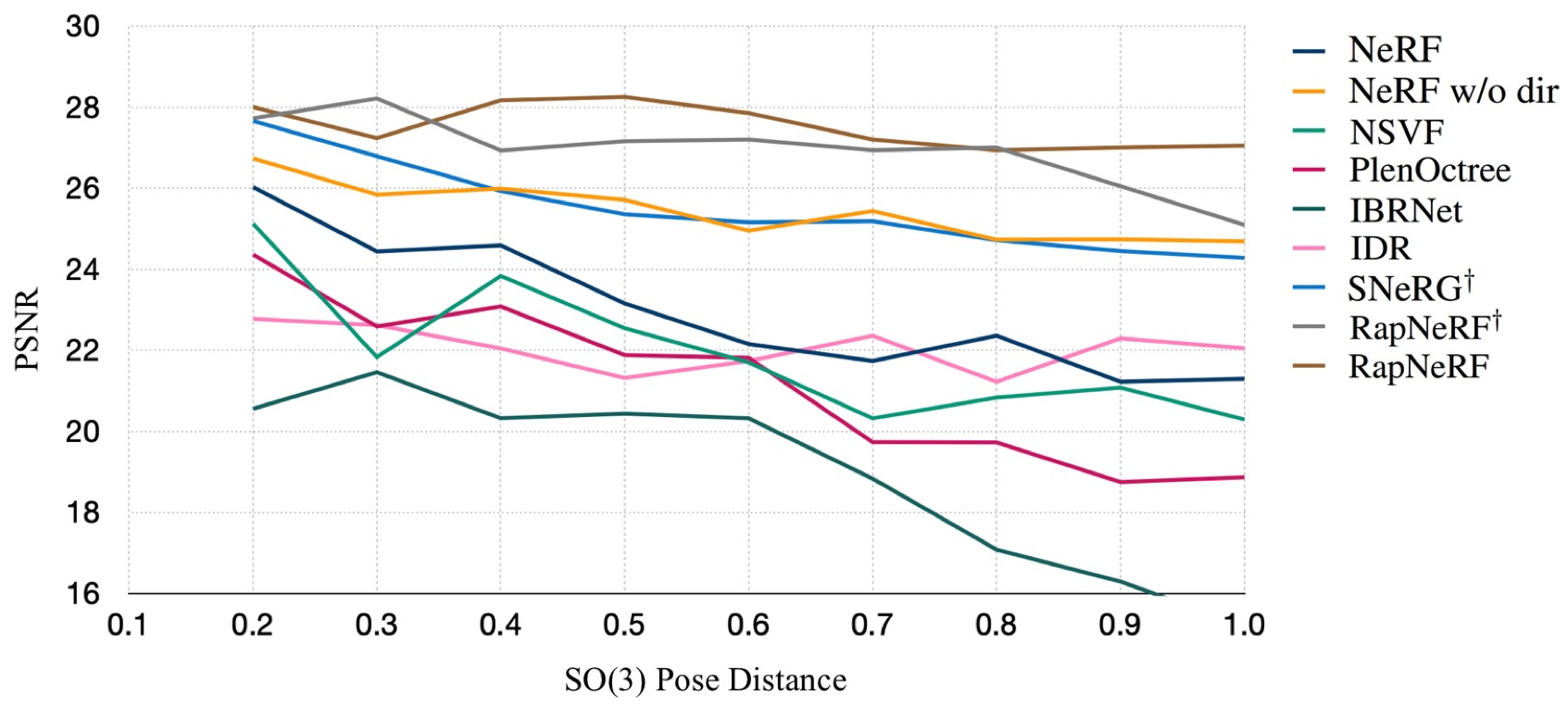}
    \caption{{\textbf{PSNR vs. SO(3) Pose Distance.} We draw the curves by computing the PSNR score and SO(3) pose distance ($D_y$) for each test image in the whole MobileObject dataset. RapNeRF yields consistently superior PSNR while $D_y$ becomes larger.
    }}
    \label{fig:psnr_curve}
\end{figure}

\begin{figure*}[t!]
    \centering
    \includegraphics[width=1.0\textwidth]{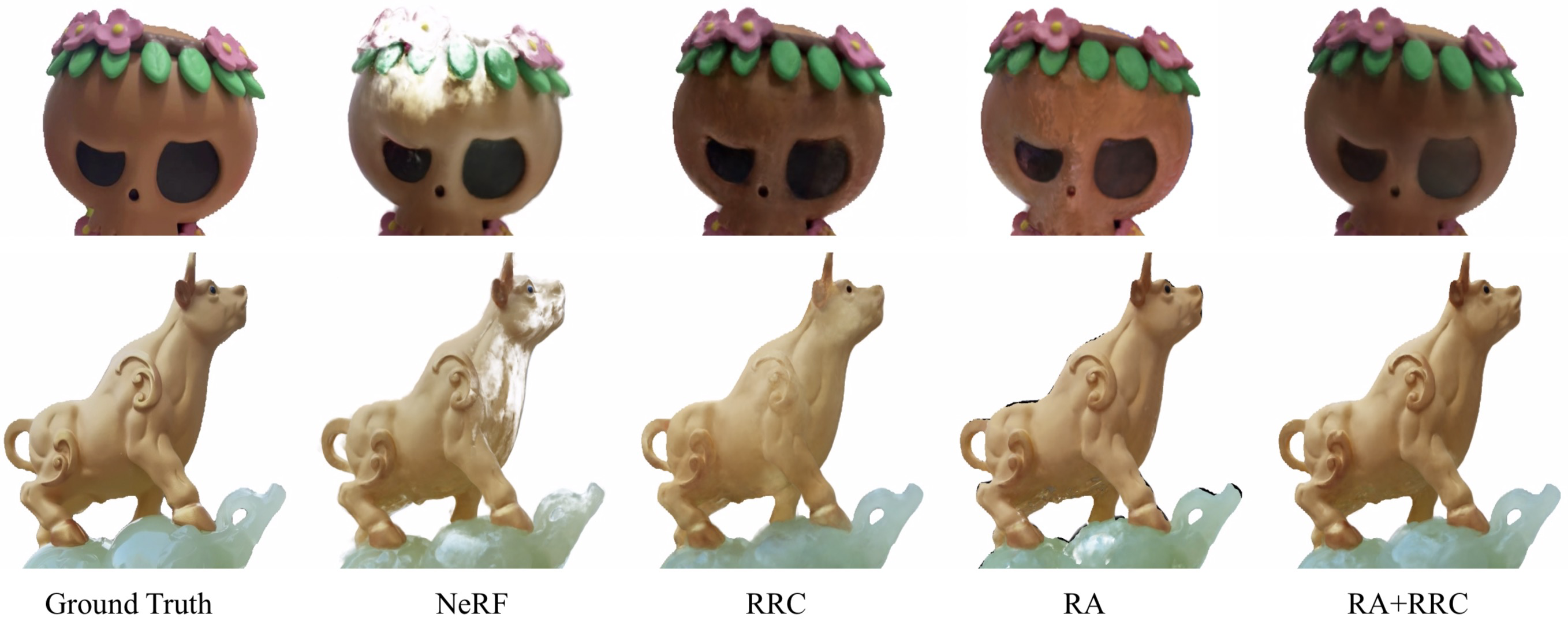}
    \caption{{We visualize the impacts of the proposed random ray casting (RRC) policy and ray atlas (RA). We can find that (1) fine-tuning NeRF with either RRC or RA could well remedy the artifact issue; and (2) RA and RRC are compatible with each other. 
    }}
    \label{fig:rrc_rm}
\end{figure*}

\noindent \textbf{Ray Priors.} We study the effectiveness of the core components in RapNeRF, \emph{i.e.}, random ray casting (RRC) and ray atlas (RA), for novel view extrapolation. As depicted in Table \ref{tab:rap}, while RA yields a PSNR improvement of 1.24 over the NeRF baseline, RRC outperforms NeRF (24.05) by a remarkable margin ($+3.5$ on PSNR). Moreover, the complete model obtains a further performance gain and produces a significant PSNR of 28.90. Some qualitative studies are presented in Figure~\ref{fig:rrc_rm}. We find, in comparisons with RA, RRC bakes more lighting effects, but it would generate smoother renderings under extrapolated viewpoints. RA would produce artifacts such as slight bump surfaces and white noises. RRC can remedy issues caused by RA.
\newline

\noindent \textbf{Pose Distance.} We study the robustness of our method to pose distance. We first present the ``PSNR vs. SO(3) pose distance'' curves obtained by different SOTA approaches in Figure~\ref{fig:psnr_curve}. Here, we draw the curves by computing the PSNR score and SO(3) pose distance ($D_y$) for each test image in the whole MobileObject dataset. We also make a comparison with NeRF using different test splits in Table~\ref{tab:vd}.
Specifically, for each object in MobileObject, we combine its test and discarded images (labeled in ``Blue'' and ``Gray'' in Figure~\ref{fig:setting}). Then we calculate the distance between these images and the training set based on Eqn.~\ref{eqn:pose-distance} and resplit them into three disjoint sets.
From Figure~\ref{fig:psnr_curve} and Table~\ref{tab:vd}, the performance of the compared approaches decreases drastically as $D_y$ becoming larger. It is worth mentioning that, for the ``Close'' setting, there are still some shifts between test viewpoints and training viewpoints (like the ``Gray'' and ``Red'' examples). Thereby, it is not surprising that NeRF can only reach 26.23 on PSNR. Luckily, RapNeRF yields consistently promising PSNR even for the ``Far'' setting. RapNeRF could generate a visually appealing rendering in these challenging cases, but other methods produce severe artifacts.
\newline

\noindent \textbf{$\eta$ in RRC.} Our random ray casting (RRC) policy allows pseudo-label assigning for randomly generated virtual rays during the training process. We introduce azimuth randomness $\Delta\varphi$ and elevation randomness $\Delta\theta$ to the ray vectors in spherical space. Here, $\Delta\varphi$ and $\Delta\theta$ are uniformly sampled from a pre-defined interval $[-\eta, \eta]$. We show quantitative results on \textit{Skull} with different elevation threshold $\eta$ for $\Delta\theta$ in Table \ref{tab:theta}. The selection $\eta = 30^{\circ}$ achieves the best performance. The image quality shows a slight decrease when $\eta$ is greater than $30^{\circ}$. A possible reason is that RRC does not consider the self-occlusions of objects.
\newline

\setlength\tabcolsep{4pt}
\begin{table}[t]
\centering

\begin{tabular}{c c c c c c c}
\toprule
 $\eta$ & $10^{\circ}$ & $20^{\circ}$ & $30^{\circ}$ & $40^{\circ}$ & $50^{\circ}$ & $60^{\circ}$ \\
\midrule\midrule
PSNR $\uparrow$ & 26.89 & 27.32 & 27.86 & 26.72 & 27.38 & 27.29\\
SSIM $\uparrow$ & 0.963 & 0.965 & 0.967 & 0.964 & 0.966 & 0.965\\
LPIPS$\downarrow$ & 0.036 & 0.032 & 0.029 & 0.030 & 0.029 & 0.029\\

\bottomrule

\end{tabular}
\caption{Quantitative results on \textit{Skull} with threshold $\eta$ in RRC.}
\label{tab:theta}
\end{table}

\setlength\tabcolsep{6pt}
\begin{table}[t!]
\centering
\begin{tabular}{ c c c c }
\toprule
 Method  & PSNR $\uparrow$ & SSIM $\uparrow$ & LPIPS $\downarrow$ \\
\midrule\midrule
NeRF-TF \cite{mildenhall2020nerf} & 31.01 & 0.947 & 0.081  \\
NeRF-TF (w/o dir) & 27.66 & 0.925 & 0.117  \\
\midrule
JaxNeRF \cite{jaxnerf2020github} & 31.65 & 0.952 & 0.051  \\
RapNeRF-PL & 30.08 & 0.949 & 0.067 \\
\midrule
SNeRG-Jax \cite{hedman2021baking} & 30.47 & 0.951 & 0.049  \\
RapNeRF$^\dag$-PL & 31.29 & 0.951 & 0.055 \\
\bottomrule
\end{tabular}
\caption{\textbf{Novel View Interpolation.} JaxNeRF is the Jax NeRF implementation. NeRF-TF is the official tensorflow NeRF implementation. RapNeRF is implemented via Pytorch-Lightning.}
\label{tab:sup}
\end{table}


\setlength\tabcolsep{6pt}
\begin{table*}[t!]
\centering
\begin{tabular}{ c c c c c c c c c c}
\toprule
 & \multicolumn{3}{c}{Close} & \multicolumn{3}{c}{Middle} & \multicolumn{3}{c}{Far} \\
 \cmidrule(lr){2-4}\cmidrule(lr){5-7}\cmidrule(lr){8-10}
 Method & PSNR $\uparrow$ & SSIM $\uparrow$ & LPIPS $\downarrow$ & PSNR $\uparrow$ & SSIM $\uparrow$ & LPIPS $\downarrow$ & PSNR $\uparrow$ & SSIM $\uparrow$ & LPIPS $\downarrow$ \\
\midrule\midrule
NeRF \cite{mildenhall2020nerf} & 26.23 & 0.960 & 0.075 & 23.42 & 0.944 & 0.095 & 22.51 & 0.939 & 0.099 \\
NeRF w/o dir \cite{liu2020neural} & 27.46 & 0.958 & 0.047 & 26.91 & 0.955 & 0.050 & 25.71 & 0.938 & 0.603 \\
RapNeRF$^{\dag}$ & 30.24 & 0.968 & 0.041 & 28.31 & 0.958 & 0.049 & 27.42 & 0.955 & 0.052 \\ 
RapNeRF  & \textbf{30.88} & \textbf{0.969} & \textbf{0.037} & \textbf{29.06} & \textbf{0.963} & \textbf{0.044} & \textbf{28.74} & \textbf{0.963} & \textbf{0.045} \\
\bottomrule
\end{tabular}
\caption{\textbf{Viewpoints Distance.} We further split the test and discarded images (labeled in ``Blue'' and ``Gray'') of each object into three disjoint subsets based on Eqn.~\ref{eqn:pose-distance}. The performance of NeRF \cite{mildenhall2020nerf} decreases drastically when the viewpoint distance between training and test images becomes larger. RapNeRF yields consistently promising scores for each setting.}
\label{tab:vd}
\end{table*}

\section{Limitation}
\label{sec:limitation}
As our key insight is the multi-view consistency property, RapNeRF would sacrifice view-dependent effects for novel view interpolation to secure much better performance for novel view extrapolation. We argue it is acceptable for applications such as 6-DOF immersive viewing and next-generation AR/VR systems. Specifically, when showcasing an object, a person prefers not to capture it in very high-frequency lighting conditions. Furthermore, studies like relighting might take some inspiration from RapNeRF since they need to decompose view-dependent effects from objects' base colors. 

For pure view synthesis, we provide a remedy to the slightly degenerated view-dependent effects by exploiting the deferred NeRF architecture in SNeRG \cite{hedman2021baking}. In particular, after the NeRF pre-training stage, we add a tiny MLP that maps from a 3D point's geometry feature $\text{f}$ (as explained in Sec.~\ref{subsec:NeRF}) and its corresponding direction embedding to a specular color. We remove the direction embedding in RRC and force RapNeRF only to learn the diffuse color. The color (or radiance) for a 3D point is now computed by the addition of its diffuse color and specular color. We denote this variant as RapNeRF$^{\dag}$. As reported in Table~\ref{tab:bc}, RapNeRF$^{\dag}$ imposes a quality loss of about 1dB in comparison with RapNeRF, but still outperforms other methods by a large margin on MobileObject (real scenes). Other results are presented in Table~\ref{tab:vd} and Figure~\ref{fig:psnr_curve}. Furthermore, RapNeRF$^{\dag}$ can be seen as training a deferred NeRF variant \cite{hedman2021baking} with RRC and RA. It further shows RRC and RA can be flexibly integrated into other NeRFs to improve their novel view exploration ability.

In Table~\ref{tab:sup}, we evaluate RapNeRF and RapNeRF$^\dag$ on the standard NeRF blender dataset. RapNeRF imposes a quality loss of 1.6dB compared to JaxNeRF. RapNeRF may fake view-dependent effects by hiding some reflected content inside the objects' surface as analyzed in \cite{hedman2021baking}. Moreover, RapNeRF$^\dag$ does not degrade the performance of SNeRG for novel view interpolation.






\section{Conclusion}
In this paper, we study \emph{Neural Radiance Fields} (NeRF) for novel view extrapolation where the test viewpoints are significantly different from the training viewpoints. We find that NeRF \cite{mildenhall2020nerf} often produces low-quality renderings with many artifacts under extrapolated viewpoints, even the training images can well describe the scenes. We take inspiration from the insight that the inherent appearances of a 3D surface's arbitrary visible projections should be consistent, and proposes RapNeRF as a solution. It is empowered by random ray casting (RRC) and ray atlas (RA). The former allows pseudo supervision for unseen views in an online manner, and the latter prudently rethinks the tradeoff between strong view-dependent effects and multi-view consistent renderings. We reconstruct Synthetic-NeRF \cite{mildenhall2020nerf} for our studied setting and build a MobileObject dataset that contains eight objects with real images. The comparisons with NeRF and its recent variants demonstrate the superiority of RapNeRF for novel view extrapolation. In the future, we would like to study neural radiance fields reconstruction from sparse multi-view images.   


\section*{Acknowledgement}
This work was partially supported by Alibaba Group through Alibaba Innovative Research Program. This work is done when Yuanqing Zhang is an intern at Alibaba Group.

{\small
\bibliographystyle{ieee_fullname}
\bibliography{egbib}
}

\end{document}